\documentclass{article}
\usepackage[final]{nips_2017}
\setcitestyle{numbers}
\setcitestyle{square}

\usepackage[utf8]{inputenc} 
\usepackage[T1]{fontenc}    
\usepackage{hyperref}       
\usepackage{url}            
\usepackage{booktabs}       
\usepackage{amsfonts}       
\usepackage{nicefrac}       
\usepackage{microtype}      
\usepackage{graphicx}
\usepackage{tabu}
\usepackage{verbatim}

\title{Revisiting Simple Neural Networks for Learning Representations of Knowledge Graphs}

\author{
 Srinivas Ravishankar\thanks{Research carried out during an internship at the Indian Institute of Science, Bangalore.} \\
  Department of Computer Science\\
 	R.V. College of Engineering, Bangalore\\
  \texttt{srini.shank@gmail.com} \\
  \And
  Chandrahas \\
  Department of Computer Science and Automation\\
  Indian Institute of Science, Bangalore\\
  \texttt{dewangan.chandrahas@gmail.com} \\
  \And
  Partha Pratim Talukdar \\
  Department of Computational and Data Sciences\\
  Indian Institute of Science, Bangalore\\
  \texttt{partha@talukdar.net} \\
  }

\begin{document}
\maketitle
\begin{abstract}
We address the problem of learning vector representations for entities and relations in Knowledge Graphs (KGs) for Knowledge Base Completion (KBC).
This problem has received significant attention in the past few years and multiple methods have been proposed.
Most of the existing methods in the literature use a predefined  characteristic scoring function for evaluating the correctness of KG triples.
These scoring functions distinguish correct triples (high score) from incorrect ones (low score).
However, their performance vary across different datasets.
In this work, we demonstrate that a simple neural network based score function can consistently achieve near start-of-the-art performance on multiple datasets. We also quantitatively demonstrate biases in standard benchmark datasets, and highlight the need to perform evaluation spanning various datasets. 

\end{abstract}

\maketitle

\section{Introduction}

Knowledge Graphs (KGs) such as 
NELL \cite{mitchell2015never} and Freebase\cite{bollacker2008freebase} are repositories of information stored as multi-relational graphs.
They are used in many applications such as information extraction, question answering etc.
oSuch KGs contain world knowledge in the form of relational triples $(h, r, t)$, where entity $h$ is connected to entity $t$ using directed relation $r$. 
For example, \textit{(DonaldTrump, PresidentOf, USA)} would indicate the fact that \textit{Donald Trump} is the president of \textit{USA}.
Although current KGs are fairly large, containing millions of facts, they tend to be quite sparse \cite{west2014knowledge}.
To overcome this sparsity, Knowledge Base Completion (KBC) or Link Prediction is performed to infer missing facts from existing ones. Low dimensional vector representations of entities and relations, also called embeddings, have been extensively used for this problem \cite{trouillon2016complex,nickel2016holographic}.
Most of these methods use a characteristic score function which distinguishes correct triples (high score) from incorrect triples (low score).
Some of these methods and their scoring functions are summarized in Table \ref{tab:scoreFunctions}.

\begin{table}[t]%
\begin{minipage}{\columnwidth}
\begin{center}
\begin{tabular}{clclc}
  \toprule
  TransE \cite{bordes2013translating}    & $- ||\textbf{h} + \textbf{r} - \textbf{t}||$\\
  \hline
  HolE \cite{nickel2016holographic}     & $(\textbf{r}^{T} . (\textbf{h} \star \textbf{t}))$\\
  \hline
  DistMult \cite{yang2014embedding}  & $<\textbf{h}, \textbf{r}, \textbf{t}>$\\
  \hline
  ComplEx \cite{trouillon2016complex}   & $Re(<\textbf{h}, \textbf{r}, \overline{\textbf{t}}>)$\\
  \bottomrule
\end{tabular}
\end{center}
\centering
\end{minipage}
\caption{\label{tab:scoreFunctions}Score functions of some well known Knowledge Graph embedding models. Here \textbf{h}, \textbf{t}, \textbf{r} are vector embeddings for entities h, t and relation r, respectively. $\star$, $<\cdot,\cdot,\cdot>$ and $Re$ represent circular correlation, sum of component-wise product, and real part of complex number, respectively.}
\end{table}%

WN18 and FB15k are two standard benchmarks datasets for evaluating link prediction over KGs. Previous research have shown that these two datasets suffer from \emph{inverse relation bias}  \cite{2017arXiv170701476D} (more in Section \ref{sec:bias}). Performance in these datasets is largely dependent on the model's ability to predict inverse relations, at the expense of other independent relations. In fact, a simple rule-based model exploiting such bias was shown to have achieved state-of-the-art performance in these two datasets \cite{2017arXiv170701476D}. To overcome this shortcoming, several variants, such as FB15k-237  \cite{toutanova2015representing}, WN18RR \cite{trouillon2016complex}, FB13 and WN11 \cite{socher2013reasoning} have been proposed in the literature. 

ComplEx \cite{trouillon2016complex} and HolE \cite{nickel2016holographic} are two popular KG embedding techniques which achieve state-of-the-art performance on the WN18 and FB15k datasets. However, we observe that such methods do not perform as well uniformly across all the other datasets mentioned above. This may suggest that using a predefined scoring function, as in ComplEx and HolE, might not be the best option to achieve competitive results on all datasets. Ideally, we would prefer a model that achieves near state-of-the-art performance on any given dataset.

In this paper, we demonstrate that a simple neural network based score function that can adapt to different datasets and achieve near state-of-the-art performance on multiple datasets.
The main contributions of this papers can be summarized as follows.
\begin{itemize}
\item We quantitatively demonstrate the severity of the inverse relation bias in the standard benchmark datasets.
\item We empirically show that current state-of-the-art methods do not perform consistently well over different datasets.
\item We demonstrate that ER-MLP\cite{dong2014knowledge}, a simple neural network based scoring function, has the ability to adapt to different datasets achieving near state-of-the-art performance consistently. We also consider a variant, ER-MLP-2d.
\end{itemize}
Code is available at \href{https://github.com/Srinivas-R/AKBC-2017-Paper-14.git}{https://github.com/Srinivas-R/AKBC-2017-Paper-14.git}.

\section{Related Work}

Several methods have been proposed for learning KG embeddings. 
They differ in the way entities and relations are modeled, the score function used for scoring triples, and the loss function used for training. For example, TransE \cite{bordes2013translating} uses real vectors for representing both entities and relations, while RESCAL \cite{nickel2013logistic} uses real vectors for entities, and real matrices for relations.

{\bf Translational Models}: One of the initial models for KG embeddings is TransE \cite{bordes2013translating}, which models relation $r$ as translation vectors from head entity $h$ to tail entity $t$ for a given triple $(h,r,t)$. 
A pair-wise ranking loss is then used for learning these embeddings.
Following the basic idea of translation vectors in TransE, there have been many methods which improve the performance. Some of these methods are TransH \cite{wang2014knowledge}, TransR \cite{lin2015learning}, TransA \cite{xiao2015transa}, TransG \cite{xiao2015transg} etc.

{\bf Multiplicative Models}: HolE \cite{nickel2016holographic} and ComplEx \cite{trouillon2016complex} are recent methods which achieve state-of-the-art performance in link prediction in commonly used datasets FB15k and WN18. HolE models entities and relations as real vectors and can handle asymmetric relations.
ComplEx uses complex vectors and can handle symmetric, asymmetric as well as anti-symmetric relations. We use these methods as representatives of the state-of-the-art in our experiments.

{\bf Neural Models}: Several methods that use neural networks for scoring triples have been proposed. 
Notable among them are NTN \cite{socher2013reasoning}, CONV \cite{toutanova2015representing}, ConvE \cite{2017arXiv170701476D}, and R-GCN \cite{2017arXiv170306103S}.
CONV uses the internal structure of textual relations as input to a Convolutional Neural Network.
NTN learns a tensor net \cite{socher2013reasoning} for each relation in the knowledge graph.
ConvE uses convolutional neural networks (CNNs) over reshaped input vectors for scoring triples.
R-GCN takes a different approach and uses Graph Convolutional Networks to obtain embeddings from the graph. DistMult (or some other linear model) is then used on these embeddings to obtain a score. 
We focus on simple neural models, such as ER-MLP \cite{2017arXiv170306103S}, and find that such simple models are more effective in KG embedding and link prediction than more complicated models such as ConvE or R-GCN. 

\section{Knowledge Graph Embedding using Simple Neural Networks}

Rather than using a predefined function to score triples, and then learn embeddings to fit this scoring function, we use a neural network to \emph{jointly} learn both the scoring function and embeddings together to fit the dataset.

\subsection{Neural Network as a Score Function}
\begin{figure}[h]
\begin{center}
  \includegraphics[scale=0.5]{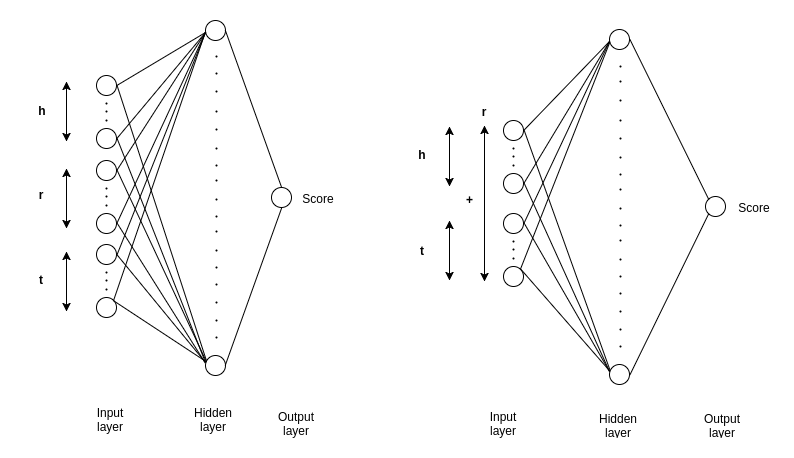}
\end{center}
  \caption{Architecture of (a) ER-MLP, (b) ER-MLP-2d}
  \label{fig:one}
\end{figure}

We use a simple feed-forward Neural Network with a single hidden layer as the approximator of the scoring function of a given triple. In particular, we use ER-MLP \cite{2017arXiv170306103S}, a previously proposed neural network model for KG embedding, and ER-MLP-2d, a variant of ER-MLP we propose. Architectures  of the two models are shown in Figure~\ref{fig:one}.

Let $\mathbf{h}$, $\mathbf{t} \in \mathbb{R}^{d}$ be the $d$-dimensional embeddings of the entities $h$ and $t$ respectively\footnote{We use boldface to refer to embeddings of corresponding italicized objects}. Similarly, $\mathbf{r}$ is the embedding of relation $r$, whose dimensions are $d$ and $2d$ in ER-MLP and ER-MLP-2d, respectively, as we shall explain below. In ER-MLP, the head, relation and tail embeddings are concatenated and fed as input to the NN, so its input layer is of size $3d$. In ER-MLP-2d, the concatenated head and tail embeddings are translated using the relation embedding of size $2d$, so input size in ER-MLP-2d is $2d$. Both models have a single fully connected hidden layer. This leads to an output node, which is taken as the score of the given triple $(h, r, t)$.

Let $g()$ denote the activation function and $[\textbf{a},\textbf{b}]$ denote concatenation of vectors $\textbf{a}$ and $\textbf{b}$. Let $M_1$ and $A_1$ be respectively the hidden and output layer weight matrices in ER-MLP. $b_1$ is a single bias value. Let the equivalent parameters in ER-MLP-2d be $M_2$, $A_2$, and $b_2$. The triple scoring functions for ER-MLP and ER-MLP-2d respectively are given below.


\begin{eqnarray*}
f_{\mathrm{ER-MLP}}(\mathbf{h},\mathbf{r},\mathbf{t}) &=& (A_1 * g(M_1 * [\mathbf{h},\mathbf{r},\mathbf{t}])) + b_1 \\
f_{\mathrm{ER-MLP-2d}}(\mathbf{h},\mathbf{r},\mathbf{t}) &=& (A_2 * g(M_2 * ([\mathbf{h},\mathbf{t}] + \mathbf{r}))) + b_2
\end{eqnarray*}

We consider the sigmoid function, $\sigma(f(\mathbf{h},\mathbf{r},\mathbf{t}))$, to be the probability of correctness of a triple. We train the model to assign probability of $1$ to correct triples and $0$ to incorrect triples. Let $\mathcal{T} = \{((h, r, t), y)\}$ be the set of positive and (sampled) negative triples, with label $y \in \{1, 0\}$. We optimize the cross-entropy loss given below, with $f$ replaced by $f_{\mathrm{ER-MLP}}$ and $f_{\mathrm{ER-MLP-2d}}$ for ER-MLP and ER-MLP-2d, respectively.
%
\begin{eqnarray*}
Loss &=& - \sum_{((h, r, t), y) \in \mathcal{T}} [y \times \log(\sigma(f(\mathbf{h},\mathbf{r},\mathbf{t}))) + (1 - y) \times \log(1-\sigma(f(\mathbf{h},\mathbf{r},\mathbf{t})))]
\end{eqnarray*}

\section{Experimental Setup}

\subsection{Implementation Details}

For initialization of embeddings, Uniform initialization with range $[-1,1]$ was used. Xavier initialization was used for weights.
In the neural network-based models, we used Dropout \cite{srivastava2014dropout} with p = 0.5 on the hidden layer to prevent overfitting. 
The regularization parameter for weight decay was chosen from \{0.001, 0.01, 0.1\} based on cross validation.
We chose the hidden layer size between \{10d, 20d\}. Since the size of this layer determines the expressive power of the model, datasets with simple relations (lower relation-specific indigree) require smaller number of hidden units and more difficult datasets require a higher number to achieve optimal performance.\newline\newline
ReLU \cite{nair2010rectified} activation is used in the hidden layer to achieve fast convergence. To minimize the objective function, we used ADAM \cite{kingma2014adam} with learning rate 0.001. Dimensionality of entity and relation embeddings were set equal in ER-MLP, i.e., $d_{e} = d_{r} = d$. For ER-MLP-2d, we have $2d_{e} = d_{r} = 2d$. $d$ was cross validated on \{100, 200\} for all datasets.
All experiments were run using Tensorflow on a single GTX 1080 GPU. To achieve maximum GPU utilization, we set the batch size larger than used previously in literature, choosing from \{10000, 20000, 50000\} using cross validation. For sampling negative triples, we used \textit{bernoulli} method as described in \cite{bordes2013translating}.

\begin{table}[t]
\centering
\begin{tabular}{cccc}
\hline
Model & Number of Parameters & WN18 & FB15K\\ \hline
HolE & $N_ed + N_rd$ & $4.096\times10^6$ & $1.629\times10^6$\\ \hline
ComplEx & $2N_ed + 2N_rd$ & $8.192\times10^6$ & $3.259\times10^6$\\ \hline
ConvE & $N_ed + N_rd$ & $4.096\times10^6$ & $1.629\times10^6$\\ \hline
ER-MLP & $N_ed + N_rd + 30d^2 + 10d$ & $4.397\times10^6$ & $1.83\times10^6$\\ \hline
ER-MLP-2d & $N_ed + 2N_rd + 20d^2 + 10d$ & $4.298\times10^6$ & $1.965\times10^6$ \\ \hline
\end{tabular}
\caption{\label{tab:memory}Number of parameters of various methods over the WN18 and FB15K datasets. Above, $d$ represents entity embedding size, $N_e$ is the number of entities, and $N_r$ is the number of relations.}
\end{table}

\subsection{Datasets}

We ran experiments on the datasets listed in Table \ref{tab:two}.
Previous work has noted that the two benchmark datasets -- FB15K and WN18 -- have a high number of redundant and reversible relations \cite{toutanova2015representing}.    A simple rule-based model, exploiting such deficiencies, was shown to have achieved state-of-the-art performance in these datasets \cite{2017arXiv170701476D}. This suggests that evaluation restricted only to these two datasets may not be an accurate indication of the model's capability. In order to address this issue, we evaluate model performance over six datasets, as summarized in Table \ref{tab:two}.

\subsubsection{Inverse Relation Bias}
\label{sec:bias}

\begin{table}[t]
\centering
\begin{tabular}{ccccccc}
\hline
& WN18    & FB15K   & WN18RR & FB15K-237 & WN11 & FB13  \\ \hline
Number of Relations            & 18      & 1345    & 11     & 237       & 11   & 13    \\ \hline
Percentage of Trivial Test Triples             & 72.12\% & 54.42\% & 0\%    & 0\%       & 0\%  & 0\%   \\ \hline
\end{tabular}
\caption{\label{tab:two}Inverse Relation Bias present in various datasets. Please see Section \ref{sec:bias} for more details.}
\end{table}

In a knowledge graph, a pair of relations r and r' are said to be inverse relations if a correct triple (h,r,t) implies the existence of another correct triple (t,r',h), and vice versa.
A \emph{trivial triple} refers to the existence of a triple $(h, r, t)$ in the \emph{test} dataset when $(t, r', h)$ is already present in the \emph{training} dataset, with $r$ and $r'$ being inverse relations. A model that can learn inverse relations well at the expense of other types of relations will still achieve very good performance on datasets involving such biased relations. This is undesirable, since our goal is to learn effective embeddings of highly multi-relational graphs.

We quantitatively investigated the bias of various datasets towards such inverse relations by measuring the fraction of trivial triples present in them. The results are summarized in Table \ref{tab:two}. Using the training dataset, each pair of relations were tested for inversion. They were identified as inverses if 80\% or more triples that contained one relation appeared as inverse triple involving the other relation. As can be seen from the table, the two standard benchmark datasets -- FB15K and WN18 -- have a large number of trivial triples. This is in contrast to four other pre-existing datasets in literature -- FB13, WN11, FB15K-237, and WN18RR -- which do not suffer from such bias. As mentioned above, we perform experiments spanning all six datasets.

\section{Experiments}
\label{sec:expts}

We chose HolE and ComplEx for comparison as these are the state-of-the-art in the current literature. Both models were re-implemented by us for fair comparison. We were able to achieve better performance for HolE on FB15K than what was reported in the original paper.  Available results of ConvE and R-GCN have been taken from \cite{2017arXiv170701476D} for comparison. We evaluated these models on the link prediction task and the results are reported in Table \ref{tab:three}.

\begin{table}[t]

\begin{center}
 \begin{tabular}{cccccccc}
 \hline
&\multicolumn{3}{c}{WN18}
 &&
 \multicolumn{3}{c}{FB15K} \\
 & Hits@10 & MR & MRR && Hits@10 & MR & MRR \\ [0.5ex] 
 \hline
 HolE & 94.12 & 810 & 0.934 && 84.35 & 113 & 0.64\\ 
 ComplEx & 94.64 & 826 & 0.938 && \textbf{87.33} & 113 & \textbf{0.75} \\
 ConvE & 95.5 & 504 & \textbf{0.942} && 87.3 & \textbf{64} & 0.745\\
 R-GCN & \textbf{96.4} & - & 0.814 && 84.2 & - & 0.696\\
 \hline
 ER-MLP & 94.2 & \textbf{299} & 0.895 && 80.14 & 81 & 0.57\\
 ER-MLP-2d & 93.66 & 372 & 0.893 && 80.04 & 81 & 0.567\\
 \hline 
\end{tabular}
\end{center}

\begin{center}
 \begin{tabular}{cccccccc}
 \hline
 &\multicolumn{3}{c}{FB15K-237}
 &&
 \multicolumn{3}{c}{WN18RR} \\
 & Hits@10 & MR & MRR && Hits@10 & MR & MRR \\ [0.5ex]
 \hline
 HolE & 47.0 & 501 & 0.298 && 42.4 & 6129 & 0.395\\ 
 ComplEx & 50.7 & 381 & 0.326 && \textbf{50.7} & 5261 & \textbf{0.444}\\
 ConvE & 45.8 & 330 & 0.301 && 41.1 & 7323 & 0.342\\
 R-GCN & 41.7 & - & 0.248 && - & - & -\\
 \hline
 ER-MLP & 54.0 & \textbf{219} & \textbf{0.342} && 41.92 & 4798 & 0.366\\
 ER-MLP-2d & \textbf{54.65} & 234 & 0.338 && 42.1 & \textbf{4233} & 0.358\\
 \hline 
\end{tabular}
\end{center}

\begin{center}
 \begin{tabular}{cccccccc}
 \hline
&\multicolumn{3}{c}{FB13}
 &&
 \multicolumn{3}{c}{WN11} \\
 & Hits@10 & MR & MRR && Hits@10 & MR & MRR \\ [0.5ex] 
 \hline
 HolE & 54.87 & 1436 & 0.392 && 10.48 & 10182 & 0.059\\ 
 ComplEx & 51.38 & 6816 & 0.382 && 10.9 & 11134 & 0.071 \\
 \hline
 ER-MLP & \textbf{63.13} & \textbf{705} & \textbf{0.479} && \textbf{14.01} & 4660 & 0.071\\
 ER-MLP-2d & 62.66 & 821 & 0.476 && 13.26 & \textbf{4290} & \textbf{0.073}\\
 \hline 
\end{tabular}
\end{center}
\caption{\label{tab:three}Link prediction performance of various methods on different datasets. Available results for ConvE and R-GCN are taken from \cite{2017arXiv170701476D}. Please see Section \ref{sec:expts} for more details.}

\end{table}

\subsection{Analysis of results}
Based on the results in in Table \ref{tab:three}, we make the following observations.
\begin{enumerate}
\item Neural network based models achieve state-of-the-art performance on WN11, FB13 and FB15K-237, and perform competitively on WN18. This is encouraging since all these datasets (except WN18) have zero trivial triples (Table \ref{tab:two}) and are therefore more challenging compared to the other datasets.
\item Surprisingly, linear models such as ComplEx and HolE perform better than neural models on WN18RR, a dataset without trivial triples. This behavior has been related with the PageRank (a measure of indegree) of central nodes in different datasets by  \citep{2017arXiv170701476D}. They found that linear models perform better on simpler datasets with low relation-specific indegree, such as WordNet. This is because they are easier to optimize and are able to find better local minima. Neural models show superior performance on complex datasets with higher relation-specific indegree.
\item Despite the effectiveness of a simple neural model like ER-MLP, such methods haven't received much attention in recent literature. Even though ER-MLP was compared against HolE in \cite{nickel2016holographic}, a rigorous comparison involving diverse datasets was missing. Results in this paper address this gap, and  show that such simple models merit further consideration in the future.
\end{enumerate}

\section{Conclusions and Future Work}
In this work, we showed that the current state-of-the-art models do not achieve uniformly good performance across different datasets, and that the current benchmark datasets can be misleading when evaluating a model's ability to represent multi-relational graphs.
We recommend that models henceforth be evaluated on multiple datasets so as to ensure their adaptability to Knowledge Graphs with different characteristics.
We also showed that a neural network with a single hidden layer, which learns the scoring function together with the embeddings, can achieve competitive performance across datasets in spite of its simplicity.
In future, we plan to identify the characteristics of datasets that determine the performance of various models.

\small
\bibliographystyle{ACM-Reference-Format}
\bibliography{bibliography}


\begin{thebibliography}{00}


\ifx \showCODEN    \undefined \def \showCODEN     #1{\unskip}     \fi
\ifx \showDOI      \undefined \def \showDOI       #1{#1}\fi
\ifx \showISBNx    \undefined \def \showISBNx     #1{\unskip}     \fi
\ifx \showISBNxiii \undefined \def \showISBNxiii  #1{\unskip}     \fi
\ifx \showISSN     \undefined \def \showISSN      #1{\unskip}     \fi
\ifx \showLCCN     \undefined \def \showLCCN      #1{\unskip}     \fi
\ifx \shownote     \undefined \def \shownote      #1{#1}          \fi
\ifx \showarticletitle \undefined \def \showarticletitle #1{#1}   \fi
\ifx \showURL      \undefined \def \showURL       {\relax}        \fi
\providecommand\bibfield[2]{#2}
\providecommand\bibinfo[2]{#2}
\providecommand\natexlab[1]{#1}
\providecommand\showeprint[2][]{arXiv:#2}

\bibitem[\protect\citeauthoryear{Bollacker, Evans, Paritosh, Sturge, and
  Taylor}{Bollacker et~al\mbox{.}}{2008}]%
        {bollacker2008freebase}
\bibfield{author}{\bibinfo{person}{Kurt Bollacker}, \bibinfo{person}{Colin
  Evans}, \bibinfo{person}{Praveen Paritosh}, \bibinfo{person}{Tim Sturge},
  {and} \bibinfo{person}{Jamie Taylor}.} \bibinfo{year}{2008}\natexlab{}.
\newblock \showarticletitle{Freebase: a collaboratively created graph database
  for structuring human knowledge}. In \bibinfo{booktitle}{{\em Proceedings of
  the 2008 ACM SIGMOD international conference on Management of data}}. AcM,
  \bibinfo{pages}{1247--1250}.
\newblock


\bibitem[\protect\citeauthoryear{Bordes, Usunier, Garcia-Duran, Weston, and
  Yakhnenko}{Bordes et~al\mbox{.}}{2013}]%
        {bordes2013translating}
\bibfield{author}{\bibinfo{person}{Antoine Bordes}, \bibinfo{person}{Nicolas
  Usunier}, \bibinfo{person}{Alberto Garcia-Duran}, \bibinfo{person}{Jason
  Weston}, {and} \bibinfo{person}{Oksana Yakhnenko}.}
  \bibinfo{year}{2013}\natexlab{}.
\newblock \showarticletitle{Translating embeddings for modeling
  multi-relational data}. In \bibinfo{booktitle}{{\em Advances in neural
  information processing systems}}. \bibinfo{pages}{2787--2795}.
\newblock


\bibitem[\protect\citeauthoryear{{Dettmers}, {Minervini}, {Stenetorp}, and
  {Riedel}}{{Dettmers} et~al\mbox{.}}{2017}]%
        {2017arXiv170701476D}
\bibfield{author}{\bibinfo{person}{T. {Dettmers}}, \bibinfo{person}{P.
  {Minervini}}, \bibinfo{person}{P. {Stenetorp}}, {and} \bibinfo{person}{S.
  {Riedel}}.} \bibinfo{year}{2017}\natexlab{}.
\newblock \showarticletitle{{Convolutional 2D Knowledge Graph Embeddings}}.
\newblock \bibinfo{journal}{{\em ArXiv e-prints\/}} (\bibinfo{date}{July}
  \bibinfo{year}{2017}).
\newblock
\showeprint[arxiv]{cs.LG/1707.01476}


\bibitem[\protect\citeauthoryear{Dong, Gabrilovich, Heitz, Horn, Lao, Murphy,
  Strohmann, Sun, and Zhang}{Dong et~al\mbox{.}}{2014}]%
        {dong2014knowledge}
\bibfield{author}{\bibinfo{person}{Xin Dong}, \bibinfo{person}{Evgeniy
  Gabrilovich}, \bibinfo{person}{Geremy Heitz}, \bibinfo{person}{Wilko Horn},
  \bibinfo{person}{Ni Lao}, \bibinfo{person}{Kevin Murphy},
  \bibinfo{person}{Thomas Strohmann}, \bibinfo{person}{Shaohua Sun}, {and}
  \bibinfo{person}{Wei Zhang}.} \bibinfo{year}{2014}\natexlab{}.
\newblock \showarticletitle{Knowledge vault: A web-scale approach to
  probabilistic knowledge fusion}. In \bibinfo{booktitle}{{\em Proceedings of
  the 20th ACM SIGKDD international conference on Knowledge discovery and data
  mining}}. ACM, \bibinfo{pages}{601--610}.
\newblock


\bibitem[\protect\citeauthoryear{Kingma and Ba}{Kingma and Ba}{2014}]%
        {kingma2014adam}
\bibfield{author}{\bibinfo{person}{Diederik Kingma} {and}
  \bibinfo{person}{Jimmy Ba}.} \bibinfo{year}{2014}\natexlab{}.
\newblock \showarticletitle{Adam: A method for stochastic optimization}.
\newblock \bibinfo{journal}{{\em arXiv preprint arXiv:1412.6980\/}}
  (\bibinfo{year}{2014}).
\newblock


\bibitem[\protect\citeauthoryear{Lin, Liu, Sun, Liu, and Zhu}{Lin
  et~al\mbox{.}}{2015}]%
        {lin2015learning}
\bibfield{author}{\bibinfo{person}{Yankai Lin}, \bibinfo{person}{Zhiyuan Liu},
  \bibinfo{person}{Maosong Sun}, \bibinfo{person}{Yang Liu}, {and}
  \bibinfo{person}{Xuan Zhu}.} \bibinfo{year}{2015}\natexlab{}.
\newblock \showarticletitle{Learning Entity and Relation Embeddings for
  Knowledge Graph Completion.}. In \bibinfo{booktitle}{{\em AAAI}}.
  \bibinfo{pages}{2181--2187}.
\newblock


\bibitem[\protect\citeauthoryear{Mitchell, Cohen, Hruschka~Jr, Talukdar,
  Betteridge, Carlson, Mishra, Gardner, Kisiel, Krishnamurthy,
  et~al\mbox{.}}{Mitchell et~al\mbox{.}}{2015}]%
        {mitchell2015never}
\bibfield{author}{\bibinfo{person}{Tom~M Mitchell}, \bibinfo{person}{William~W
  Cohen}, \bibinfo{person}{Estevam~R Hruschka~Jr},
  \bibinfo{person}{Partha~Pratim Talukdar}, \bibinfo{person}{Justin
  Betteridge}, \bibinfo{person}{Andrew Carlson}, \bibinfo{person}{Bhavana~Dalvi
  Mishra}, \bibinfo{person}{Matthew Gardner}, \bibinfo{person}{Bryan Kisiel},
  \bibinfo{person}{Jayant Krishnamurthy}, {et~al\mbox{.}}}
  \bibinfo{year}{2015}\natexlab{}.
\newblock \showarticletitle{Never Ending Learning.}. In
  \bibinfo{booktitle}{{\em AAAI}}. \bibinfo{pages}{2302--2310}.
\newblock


\bibitem[\protect\citeauthoryear{Nair and Hinton}{Nair and Hinton}{2010}]%
        {nair2010rectified}
\bibfield{author}{\bibinfo{person}{Vinod Nair} {and}
  \bibinfo{person}{Geoffrey~E Hinton}.} \bibinfo{year}{2010}\natexlab{}.
\newblock \showarticletitle{Rectified linear units improve restricted boltzmann
  machines}. In \bibinfo{booktitle}{{\em Proceedings of the 27th international
  conference on machine learning (ICML-10)}}. \bibinfo{pages}{807--814}.
\newblock


\bibitem[\protect\citeauthoryear{Nickel, Rosasco, Poggio, et~al\mbox{.}}{Nickel
  et~al\mbox{.}}{2016}]%
        {nickel2016holographic}
\bibfield{author}{\bibinfo{person}{Maximilian Nickel}, \bibinfo{person}{Lorenzo
  Rosasco}, \bibinfo{person}{Tomaso~A Poggio}, {et~al\mbox{.}}}
  \bibinfo{year}{2016}\natexlab{}.
\newblock \showarticletitle{Holographic Embeddings of Knowledge Graphs.}. In
  \bibinfo{booktitle}{{\em AAAI}}. \bibinfo{pages}{1955--1961}.
\newblock


\bibitem[\protect\citeauthoryear{Nickel and Tresp}{Nickel and Tresp}{2013}]%
        {nickel2013logistic}
\bibfield{author}{\bibinfo{person}{Maximilian Nickel} {and}
  \bibinfo{person}{Volker Tresp}.} \bibinfo{year}{2013}\natexlab{}.
\newblock \showarticletitle{Logistic tensor factorization for multi-relational
  data}.
\newblock \bibinfo{journal}{{\em arXiv preprint arXiv:1306.2084\/}}
  (\bibinfo{year}{2013}).
\newblock


\bibitem[\protect\citeauthoryear{{Schlichtkrull}, {Kipf}, {Bloem}, {van den
  Berg}, {Titov}, and {Welling}}{{Schlichtkrull} et~al\mbox{.}}{2017}]%
        {2017arXiv170306103S}
\bibfield{author}{\bibinfo{person}{M. {Schlichtkrull}}, \bibinfo{person}{T.~N.
  {Kipf}}, \bibinfo{person}{P. {Bloem}}, \bibinfo{person}{R. {van den Berg}},
  \bibinfo{person}{I. {Titov}}, {and} \bibinfo{person}{M. {Welling}}.}
  \bibinfo{year}{2017}\natexlab{}.
\newblock \showarticletitle{{Modeling Relational Data with Graph Convolutional
  Networks}}.
\newblock \bibinfo{journal}{{\em ArXiv e-prints\/}} (\bibinfo{date}{March}
  \bibinfo{year}{2017}).
\newblock
\showeprint[arxiv]{stat.ML/1703.06103}


\bibitem[\protect\citeauthoryear{Socher, Chen, Manning, and Ng}{Socher
  et~al\mbox{.}}{2013}]%
        {socher2013reasoning}
\bibfield{author}{\bibinfo{person}{Richard Socher}, \bibinfo{person}{Danqi
  Chen}, \bibinfo{person}{Christopher~D Manning}, {and} \bibinfo{person}{Andrew
  Ng}.} \bibinfo{year}{2013}\natexlab{}.
\newblock \showarticletitle{Reasoning with neural tensor networks for knowledge
  base completion}. In \bibinfo{booktitle}{{\em Advances in neural information
  processing systems}}. \bibinfo{pages}{926--934}.
\newblock


\bibitem[\protect\citeauthoryear{Srivastava, Hinton, Krizhevsky, Sutskever, and
  Salakhutdinov}{Srivastava et~al\mbox{.}}{2014}]%
        {srivastava2014dropout}
\bibfield{author}{\bibinfo{person}{Nitish Srivastava},
  \bibinfo{person}{Geoffrey~E Hinton}, \bibinfo{person}{Alex Krizhevsky},
  \bibinfo{person}{Ilya Sutskever}, {and} \bibinfo{person}{Ruslan
  Salakhutdinov}.} \bibinfo{year}{2014}\natexlab{}.
\newblock \showarticletitle{Dropout: a simple way to prevent neural networks
  from overfitting.}
\newblock \bibinfo{journal}{{\em Journal of machine learning research\/}}
  \bibinfo{volume}{15}, \bibinfo{number}{1} (\bibinfo{year}{2014}),
  \bibinfo{pages}{1929--1958}.
\newblock


\bibitem[\protect\citeauthoryear{Toutanova, Chen, Pantel, Poon, Choudhury, and
  Gamon}{Toutanova et~al\mbox{.}}{2015}]%
        {toutanova2015representing}
\bibfield{author}{\bibinfo{person}{Kristina Toutanova}, \bibinfo{person}{Danqi
  Chen}, \bibinfo{person}{Patrick Pantel}, \bibinfo{person}{Hoifung Poon},
  \bibinfo{person}{Pallavi Choudhury}, {and} \bibinfo{person}{Michael Gamon}.}
  \bibinfo{year}{2015}\natexlab{}.
\newblock \showarticletitle{Representing Text for Joint Embedding of Text and
  Knowledge Bases.}. In \bibinfo{booktitle}{{\em EMNLP}},
  Vol.~\bibinfo{volume}{15}. \bibinfo{pages}{1499--1509}.
\newblock


\bibitem[\protect\citeauthoryear{Trouillon, Welbl, Riedel, Gaussier, and
  Bouchard}{Trouillon et~al\mbox{.}}{2016}]%
        {trouillon2016complex}
\bibfield{author}{\bibinfo{person}{Th{\'e}o Trouillon},
  \bibinfo{person}{Johannes Welbl}, \bibinfo{person}{Sebastian Riedel},
  \bibinfo{person}{{\'E}ric Gaussier}, {and} \bibinfo{person}{Guillaume
  Bouchard}.} \bibinfo{year}{2016}\natexlab{}.
\newblock \showarticletitle{Complex embeddings for simple link prediction}. In
  \bibinfo{booktitle}{{\em International Conference on Machine Learning}}.
  \bibinfo{pages}{2071--2080}.
\newblock


\bibitem[\protect\citeauthoryear{Wang, Zhang, Feng, and Chen}{Wang
  et~al\mbox{.}}{2014}]%
        {wang2014knowledge}
\bibfield{author}{\bibinfo{person}{Zhen Wang}, \bibinfo{person}{Jianwen Zhang},
  \bibinfo{person}{Jianlin Feng}, {and} \bibinfo{person}{Zheng Chen}.}
  \bibinfo{year}{2014}\natexlab{}.
\newblock \showarticletitle{Knowledge Graph Embedding by Translating on
  Hyperplanes.}. In \bibinfo{booktitle}{{\em AAAI}}.
  \bibinfo{pages}{1112--1119}.
\newblock


\bibitem[\protect\citeauthoryear{West, Gabrilovich, Murphy, Sun, Gupta, and
  Lin}{West et~al\mbox{.}}{2014}]%
        {west2014knowledge}
\bibfield{author}{\bibinfo{person}{Robert West}, \bibinfo{person}{Evgeniy
  Gabrilovich}, \bibinfo{person}{Kevin Murphy}, \bibinfo{person}{Shaohua Sun},
  \bibinfo{person}{Rahul Gupta}, {and} \bibinfo{person}{Dekang Lin}.}
  \bibinfo{year}{2014}\natexlab{}.
\newblock \showarticletitle{Knowledge base completion via search-based question
  answering}. In \bibinfo{booktitle}{{\em Proceedings of the 23rd international
  conference on World wide web}}. ACM, \bibinfo{pages}{515--526}.
\newblock


\bibitem[\protect\citeauthoryear{Xiao, Huang, Hao, and Zhu}{Xiao
  et~al\mbox{.}}{2015a}]%
        {xiao2015transa}
\bibfield{author}{\bibinfo{person}{Han Xiao}, \bibinfo{person}{Minlie Huang},
  \bibinfo{person}{Yu Hao}, {and} \bibinfo{person}{Xiaoyan Zhu}.}
  \bibinfo{year}{2015}\natexlab{a}.
\newblock \showarticletitle{Transa: An adaptive approach for knowledge graph
  embedding}.
\newblock \bibinfo{journal}{{\em arXiv preprint arXiv:1509.05490\/}}
  (\bibinfo{year}{2015}).
\newblock


\bibitem[\protect\citeauthoryear{Xiao, Huang, Hao, and Zhu}{Xiao
  et~al\mbox{.}}{2015b}]%
        {xiao2015transg}
\bibfield{author}{\bibinfo{person}{Han Xiao}, \bibinfo{person}{Minlie Huang},
  \bibinfo{person}{Yu Hao}, {and} \bibinfo{person}{Xiaoyan Zhu}.}
  \bibinfo{year}{2015}\natexlab{b}.
\newblock \showarticletitle{TransG: A Generative Mixture Model for Knowledge
  Graph Embedding}.
\newblock \bibinfo{journal}{{\em arXiv preprint arXiv:1509.05488\/}}
  (\bibinfo{year}{2015}).
\newblock


\bibitem[\protect\citeauthoryear{Yang, Yih, He, Gao, and Deng}{Yang
  et~al\mbox{.}}{2014}]%
        {yang2014embedding}
\bibfield{author}{\bibinfo{person}{Bishan Yang}, \bibinfo{person}{Wen-tau Yih},
  \bibinfo{person}{Xiaodong He}, \bibinfo{person}{Jianfeng Gao}, {and}
  \bibinfo{person}{Li Deng}.} \bibinfo{year}{2014}\natexlab{}.
\newblock \showarticletitle{Embedding entities and relations for learning and
  inference in knowledge bases}.
\newblock \bibinfo{journal}{{\em arXiv preprint arXiv:1412.6575\/}}
  (\bibinfo{year}{2014}).
\newblock


\end{thebibliography}

\end{document}